\documentclass{article} 
\usepackage{iclr2022_srml_workshop,times}

\usepackage{amsmath,amsfonts,bm}

\def\eqref#1{equation~\ref{#1}}

\def\1{\bm{1}}

\DeclareMathAlphabet{\mathsfit}{\encodingdefault}{\sfdefault}{m}{sl}
\SetMathAlphabet{\mathsfit}{bold}{\encodingdefault}{\sfdefault}{bx}{n}

\newcommand{\R}{\mathbb{R}}

\DeclareMathOperator*{\argmin}{arg\,min}

\usepackage{hyperref}
\usepackage{url}
\usepackage{booktabs}       
\usepackage{amsfonts}       
\usepackage{nicefrac}       
\usepackage{microtype}      
\usepackage{xcolor}         
\usepackage{amsmath}
\usepackage[linesnumbered,ruled]{algorithm2e}
\usepackage{pifont}
\usepackage{caption}
\usepackage{subcaption}
\usepackage{etoolbox}
\usepackage{appendix}
\usepackage{dirtytalk}
\usepackage{graphicx}

\makeatletter
\newenvironment{Ualgorithm}[1][htpb]{\def\@algocf@post@ruled{\kern\interspacealgoruled\hrule  height\algoheightrule\kern3pt\relax}%
\def\@algocf@capt@ruled{above}%
\setlength\algotitleheightrule{0pt}%
\begin{algorithm}[#1]}
{\end{algorithm}}
\makeatother

\usepackage{array}
\newcolumntype{L}{>{$}l<{$}} 

\title{Fair Machine Learning \\ under Limited Demographically Labeled Data}

\author{Mustafa Safa Ozdayi \\
The University of Texas at Dallas \\
\texttt{mustafa.ozdayi@utdallas.edu} \\
\AND
Murat Kantarcioglu \\
The University of Texas at Dallas \\
\texttt{muratk@utdallas.edu} \\
\AND
Rishabh Iyer \\
The University of Texas at Dallas \\
\texttt{rishabh.iyer@utdallas.edu}
}

\iclrfinalcopy 
\begin{document}

\maketitle

\begin{abstract}
Research has shown that, machine learning models might inherit and propagate undesired social biases encoded in the data. To address this problem, fair training algorithms are developed. However, most algorithms assume we know demographic/sensitive data features such as gender and race. This assumption falls short in scenarios where collecting demographic information is not feasible due to privacy concerns, and data protection policies. A recent line of work develops fair training methods that can function without any demographic feature on the data, that are collectively referred as Rawlsian methods. Yet, we show in experiments that, Rawlsian methods tend to exhibit relatively high bias. Given this, we look at the middle ground between the previous approaches, and consider a setting where we know the demographic attributes for only a small subset of our data. In such a setting, we design fair training algorithms which exhibit both good utility, and low bias. In particular, we show that our techniques can train models to significantly outperform Rawlsian approaches even when 0.1\% of demographic attributes are available in the training data. Furthermore, our main algorithm can accommodate multiple training objectives easily. We expand our main algorithm to achieve robustness to label noise in addition to fairness in the limited demographics setting to highlight that property as well. 
\end{abstract}

\section{Introduction}
\label{sec:intro}
Researchers have illustrated that machine learning (ML) models might exhibit discriminatory behavior.
\cite{pmlr-v81-buolamwini18a} show that many of the commercial face recognition systems tend have higher rates of error on people with darker skin color.~\cite{compas} have analyzed COMPAS, a software used by some of the U.S courts to asses defendants' likelihood to reoffend. The authors unveiled that, the software overestimates the likelihood to reoffend for black defendants, and underestimates for white defendants. Finally, it was reported that Amazon had to abandon the use of its ML based recruitment tool because it was disproportionately penalizing the women candidates (\cite{amazonCV}). \\

To address such issues, many fair training algorithms are developed such as \cite{pleiss2017fairness,empiricalFairIn,metFairIn}. At its core, these algorithms try to maximize the utility of the trained models while trying to keep the bias exhibited by them minimal with respect to some bias metric(s). In the recent literature, parity-based metrics, such as equality of odds difference and average of odds difference (\cite{equalizedOdds}), are among the most popular. This is perhaps because these metrics somewhat correspond to the fairness notions put forward by the government agencies, such as the $80\%$ rule of U.S. Equal Employment Opportunity Commission.\footnote{This rule states that, companies should be hiring protected groups at a rate that is at least 80\% of that of white men (\cite{80Rule}).}\\

However, the existing parity-based fair training approaches require the demographic/sensitive attributes, such as gender and race, to be available on the data. This renders them unusable in scenarios where collecting demographic attributes from the individuals is not possible due to privacy concerns and certain data regulations, such as European Union's GDPR (\cite{EUGDPR}). As has been noted in a survey that is conducted with industry practitioners in ~\cite{holstein2019improving}, this is a significant problem that needs to be addressed before we can see the adaptation of fair training algorithms by the industry.\\

To alleviate the aforementioned problem, ~\cite{hashimoto2018fairness} and \cite{lahoti2020fairness} developed fair training algorithms that do not rely on demographic attributes at all. However, as we experimentally show in this work, these \emph{Rawlsian}\footnote{These algorithms are referred as Rawlsian because they are based on some min-max optimization. This formulation somewhat corresponds to Rawl's principle of fairness which states (among other things) \say{Social and economic inequalities are to be arranged so that they are to the \textbf{greatest benefit of the least advantaged members} of society, consistent with the just savings principle~\cite{rawls1999theory}}} algorithms tend to exhibit high bias when parity-based metrics are concerned.\\

In short, existing parity-based fair training approaches require demographic attributes on data, and this limits their usability in real-world scenarios. In contrast, Rawlsian approaches do not require any demographic attribute, but they perform relatively poor as far as parity-based bias metrics are concerned. Given these observations, we consider the middle ground between the previous approaches, and consider a setting where we know the demographic attributes for only a small subset of our data (henceforth, we refer this setting as limited demographics setting). \emph{In summary, our goal is to develop fair training algorithms which perform good with respect to parity-based bias metrics in the limited demographics setting.} As suggested in ~\cite{holstein2019improving}, limited demographics setting is realistic for many scenarios. This is because it is likely that the trained models will be validated before they are released to the wild, and validation requires data with demographic information as long as parity-based metrics are concerned\footnote{For example, in general, creditors may not request or collect information about an applicant’s race, color, religion, national origin, or sex.
\textbf{Exceptions to this rule generally involve situations in which the information is necessary to test for compliance with fair lending rules.} [CFBP Consumer Law and Regulations, 12 CFR §1002.5]}.\\

Concretely, we show and evaluate two ways to do fair training in the limited demographics setting. We start with considering a strawman solution that imputes unknown demographic attributes. In other words, we first train a predictor model for demographic attributes on the demographically labeled portion of the data, and then predict and fill the demographic attributes for the rest of the data. Then, any existing parity-based fair training algorithm can be applied. By using this approach, we illustrate the performance of some of the existing parity-based algorithms in the limited demographics setting. Our results suggest that, even if we have only a few dozen demographically labeled data at hand, \emph{the strawman solution tends to outperform the state-of-the-art Rawlsian algorithm} for parity-based bias metrics.\\

After evaluating the strawman solution, we improve upon it by developing a novel fair training algorithm that is particularly suited to limited demographics setting. By experiments, we show that our algorithm degrades more gracefully than the strawman solution, and exhibits lower bias as the size of demographically labeled data gets smaller. Furthermore, flexibility of our formulation allows us to expand our algorithm to accommodate other training objectives in addition to fairness. For example, we know that real-world data, and sometimes even carefully curated benchmarks such as ImageNet~(\cite{deng2009imagenet}) and Cifar10~(\cite{cifar10}), might contain label noise as reported by~\cite{northcutt2021labelerrors}. 
Given this, we expand our algorithm to achieve robustness to label noise in addition to the fairness in the limited demographics setting.

We summarize our contributions below. 
\begin{itemize}
    \item  We adapt some of the existing parity-based fair training algorithms to limited demographics setting with a strawman solution, and compare the strawman solution with state-of-the-art Rawlsian method of~\cite{lahoti2020fairness}. We show that, the strawman solution tends to exhibit lower bias even when there is only a few dozen demographically labeled data at hand. 
    \item  We develop a novel fair training algorithm, based on bilevel optimization, named \emph{BiFair}, that is particularly suited to limited demographics setting. Our experiments show that BiFair tends to exhibit lower bias compared to the strawman solution as the size of the demographically labeled data gets smaller.
    \item Our formulation is flexible, and can be easily expanded to contain multiple training objectives. To illustrate the flexibility of our formalization, we extend BiFair to be robust against noise in the labels. This gives us a fair training algorithm that is robust to noisy labels in the limited demographics setting.
\end{itemize}

We organize the rest of our paper as follows: in Section~\ref{sec:background}, we provide the necessary background to the reader, and discuss the related work. In Section~\ref{sec:bifair}, we present the BiFair algorithm. In Section~\ref{sec:exps}, we present our experiments. In Section~\ref{sec:discuss}, we discuss some aspects of our work as well as provide avenues for further research. Finally in Section~\ref{sec:conclusion}, we recap our work, and conclude the paper.

\section{Background and Related Work}
\label{sec:background}

\subsection{Fairness in ML}
Fairness is a multifaceted concept that has different definitions based on context it is considered. Our main focus in this work is supervised classification, and in this domain, parity-based fairness definitions, such as statistical parity, equalized odds and equality of opportunity of~\cite{eqzOdds}, are the most prominent in the recent literature. For example, AIF360 of~\cite{aif360}, a popular toolkit for fairness research, benchmarks its results by using parity-based bias metrics. \\

Algorithms that train fair models are typically grouped under three categories: pre-processing, in-processing, and post-processing. Pre-processing methods are applied to the data prior to the training. The goal is to transform the training data in a way such that, when a model is later trained on the transformed data, it exhibits good fairness performance~(\cite{kamiran, dispImpactPre, learnFairPre, optimPre}). The in-processing methods are applied at the training time, for example by adding regularization terms or encoding hard constraints on the training objective~(\cite{prjRemover, bechavod2017learning, advIn, reductIn, metFairIn, empiricalFairIn}). Finally, post-processing methods are applied to an already trained model. They try to limit the bias of the model by adjusting the model's outputs directly, such as by negating its output on certain inputs~(\cite{roc, pleiss2017fairness, equalizedOdds}). The particular algorithm we develop, BiFair, falls into the category of in-processing methods.\\

Our work differs from the existing fairness literature primarily by the setting which we consider. As mentioned before, most of the previous work assume the existence of demographic attributes on all of the data. This assumption might not be realistic due to privacy concerns and data regulations. On the other hand, some works develop fair training algorithms that can function without any demographic attribute~(\cite{hashimoto2018fairness,lahoti2020fairness}), but we show these algorithms fall short in performance when parity-based bias metrics are concerned. In contrast, we develop approaches that perform well for parity-based metrics with limited demographic data. \\

It is also worth noting the work of ~\cite{roh2020fr}, which develops an algorithm that can train both fair, and robust models. Like this work, we also consider robustness and fairness together, but we do so as an extension of our main contribution. Yet again, our work differs from this work by the limited demographics setting we consider as well.

\subsection{Bilevel Optimization}
A bilevel optimization is type a nested optimization, where optimality of an \emph{outer} problem is subject to the optimality of an \emph{inner} problem. A general formulation of the bilevel optimization is given below,
\begin{equation*}
\min_{x, y} f(x, y^\star) \text{ subject to } y^\star \in \argmin_y g(x, y).
\end{equation*}
Here, $f:\R^n \times \R^m \rightarrow \R$ is referred as the outer problem, and $g:\R^n \times \R^m \rightarrow \R$ is referred as the inner problem. It is important to note the dependence of the outer problem to the inner problem. Due to this, one cannot solve inner and outer problems simultaneously, but rather, inner problem has to be solved first before one can treat the outer problem. \\

Recent years have seen the successful application of bilevel optimization to various areas in ML, such as meta-learning ~(\cite{finn2017model, nichol2018first, rajeswaran2019meta}), or scalable high-dimensional hyperparameter optimization~(\cite{lorraine2020optimizing, franceschi2018bilevel}). From the recent works that use bilevel optimization, the most similar works to ours are due to~\cite{ren2018learning, jenni2018deep, roh2020fairbatch}. In all of these work, bilevel optimization is used to learn a set of weights on the training dataset as in our work. In~\cite{ren2018learning}, the learned weights ensure good training under datasets with severe label imbalance, and/or noisy labels. In~\cite{jenni2018deep}, weights are learned in a way to ensure good generalization of the trained model. Finally, ~\cite{roh2020fairbatch} uses bilevel optimization to develop a fair training algorithm as in our work, but their algorithm requires that all the training data to have demographic information, where we consider a setting with limited demographic data in this work as mentioned before.

\section{Fair Training with Limited Demographic Data}
\label{sec:bifair}
In this section, we tackle the problem of training fair models when we know the demographic attributes for only a subset of the data. To this end, we first provide a strawman solution that adapts existing fair training algorithms that require demographic information on all the dataset. It turns out that, even this approach works fairly well against Rawlsian approaches as far as parity-based metrics are concerned as we experimentally show later. Then, we develop a fair training algorithm that is particularly suited to limited demographics setting by leveraging the bilevel optimization framework. Our experiments in Section~\ref{sec:exps} shows this approach degrades more gracefully than the strawman approach as the size of the demographically labeled data gets smaller. 

\subsection{Problem Setting and Notation}
We consider the problem of binary supervised classification for a dataset $\{X, y, s\}_{i=1}^n$ with a model $M_\theta$ parameterized by $\theta \in R^d$. Here, $X^{(i)} \in \R^m$ is the set of features, $y^{(i)} \in \{0, 1\}$ is the label, and $s^{(i)} \in \{0, 1\}$ is the demographic/sensitive feature (e.g., race) of the $i$th sample. A sample is referred as \emph{favorable} if its label is $1$, and \emph{unfavorable} otherwise. Similarly, the samples with a sensitive attribute of $1$ are referred as \emph{privileged}, and \emph{unprivileged} otherwise. By definition, privileged samples appear more frequently with the favorable label in the overall population,  e.g., it could be that 2/3 of favorable labels belong to the privileged group, where as this figure is 1/3 for unprivileged samples. 

Crucially, we assume that we know $s_i$ values only a for a subset of our dataset. Our goal is to learn a model can predict the labels of new samples with high accuracy, while exhibiting low bias value with respect to a parity-based bias metric(s), such as those defined in Table~\ref{tab:biasMetrics}.

\begin{table}[t]
\Large
\centering
\caption{Various parity-based bias metrics that are used to quantify the bias exhibited by models, and corresponding loss functions that can be plugged into BiFair (see Equation~\ref{eqn:bifair}). Lower values for bias metrics indicate fairer models. As can be seen, many parity-based bias metrics can simply be expressed as utility loss differences across groups, and yield differentiable loss functions.}
\label{tab:biasMetrics}

\resizebox{\textwidth}{!}{%
\begin{tabular}{ c c c }
\toprule
Metric & Definition & Fairness Loss  \\ 
\midrule

Statistical Parity Difference (SPD) & 
$|\text{PPR}^{p} - \text{PPR}^{up}|$ & 
$|L_{u|1}^{p} - L_{u|1}^{up}|$ \\

Equality of Opportunity Difference (EOD) & $|\text{TPR}^p - \text{TPR}^{up}|$ & 
 $|L_{u}^{p, fav} - L_{u}^{up, fav}|$ \\
 
Average Odds Difference (AOD) & $0.5\cdot (|\text{FPR}^p - \text{FPR}^{up}| +   |\text{TPR}^p - \text{TPR}^{up}|)$ & 
$|L_{u}^{p, unfav} - L_{u}^{up, unfav}| +  |L_{u}^{p, fav} - L_{u}^{up, unfav}|$ \\

\bottomrule
\end{tabular}%
}
\caption*{PPR, FPR, and TPR denote the positive predictive rate, false positive rate, and true positive rate, respectively.  $L_u$ denotes the utility loss function we use to train the model, such as cross-entropy loss for logistic regression and neural networks. The superscripts \emph{p}, and \emph{up} denote the privileged, and unprivileged groups. Similarly, the superscripts \emph{fav} and \emph{unfav} denote the favorable, and the unfavorable label. For example, $L_{u}^{p, fav}$ denotes the average utility loss computed over privileged and favorable samples. Note that, in contrast to EOD and AOD, SPD does not take ground-truth labels of inputs into account. Therefore, we have to compute the utility loss by setting the target as $1$ for SPD, denoted as $L_{u|1}$, regardless of the actual label of the data instance.}
\end{table}

\subsection{A Strawman Solution}
\label{subsec:strawman}
Let $D_{tr}$ denote our training dataset, and let $D_{dem}$ be the subset of $D_{tr}$ whose demographic attributes are known. Then, we can adapt any existing parity-based fair training algorithm to limited demographics setting. First, (i) train a predictor model for demographic features using $D_{dem}$, then, (ii) fill the demographic features for the remaining of training data by having the model predict demographic attributes from other features, and finally, (iii) run the fair training algorithm on $D_{tr}$ as usual. \\

Although this data imputation method is pretty straightforward, there are few points to discuss. First, we can see that the performance of this approach is essentially determined by how accurately we can predict demographic features from the other features. So when using this approach, we are implicitly assuming that there exists a correlation between the sensitive feature ($s_i$), and other features ($X_i$). In our experiments, we have observed this assumption generally holds. Regardless, this correlation can change from dataset to dataset, and might affect the stability of the strawman approach. Therefore, it is worth to note that the next algorithm we develop makes no such assumption.

\subsection{BiFair}
We now describe a fair training algorithm that is particularly suited to limited demographics setting, named \emph{BiFair}, by leveraging the bilevel optimization framework. Briefly, we introduce a set of weights $w$ on the training dataset, such that training on the weighted dataset yields both good utility, and low bias for the model. We learn the values of $w$ concurrent to the model training by solving a bilevel optimization problem. Concretely, let $L_u$ be a loss function that we use to train the model. For example, $L_u$ could be the hinge loss if our model is a SVM, or it could be cross-entropy if it is a neural network. Further, let $L_f$ be differentiable fairness loss that is associated with a bias metric, such that, by minimizing $L_f$, we can reduce the bias of the model. Then, we can formulate our learning objective as follows,

\begin{equation}
\label{eqn:bifair}
\begin{split}
w^*, \theta^* \in &\argmin_{w, \theta} L_f(M_{\theta^\star}, D_{dem}), \\
  &\text{subject to } \theta^\star \in \argmin_{\theta} \sum_i w^{(i)} \cdot L_u(M_{\theta}, D_{tr}^{(i)}).
\end{split}
\end{equation}
As is seen, in the inner optimization, we minimize $L_u$ on the weighted training dataset. In the outer optimization, the weights are adjusted to minimize the fairness loss on the portion of the dataset where we have access to the demographic attributes (hence, we can compute any of the fairness losses given in Table~\ref{tab:biasMetrics} on that portion).\\

Many models of practical interest yields no closed-form solution to the inner problem, but rather, are optimized by iterative methods, e.g., by gradient descent. So, finding an optimal solution to the inner problem formulated in Equation~\ref{eqn:bifair} is usually a costly process. One workaround of this, is to relax the inner problem by finding an approximate solution to it as presented in~\cite{domke}. Briefly, we can approximate the solution to the inner problem by taking a few steps of gradient descent, and then compute the gradient for the outer problem at the approximated solution. We can then update the training data weights using the gradient of the outer problem, and repeat this  until the outer level problem converges. This gives us an algorithm that is straightforward to implement with ML frameworks that provide automatic differentiation such as PyTorch~(\cite{PyTorch}) and TensorFlow~(\cite{tensorflow2015-whitepaper}). The pseudo-code of our algorithm is presented in Algorithm~\ref{alg:biFair}. Briefly, the lines 3-9 correspond to approximating the inner problem, and in line 14 and 15, we compute the gradient for the outer problem, and update the dataset weights, respectively.

\subsection{Extending BiFair for Robustness}
\label{subsec:robustness}
As mentioned before in Section~\ref{sec:intro}, the bilevel optimization is flexible in the sense that, we can target multiple objectives at the same time by minimal change in the formulation. To highlight this, we consider fairness and robustness together. In what follows, we assume $D_{tr}$ might contain noisy labels, and $D_{dem}$ has clean labels. With such an assumption, we can add robustness to BiFair by simply adding the utility loss to the outer-level problem. That is, the new formulation becomes,

\begin{equation}
\label{eqn:bifairRobust}
\begin{split}
w^*, \theta^* \in &\argmin_{w, \theta} L_f(M_{\theta^\star}, D_{dem}) + \lambda \cdot L_u(M_{\theta^\star},  D_{dem}), \\
  &\text{subject to } \theta^\star \in \argmin_{\theta} \sum_i w^{(i)} \cdot L_u(M_{\theta}, D_{tr}^{(i)}).
\end{split}
\end{equation}
 where $\lambda \geq 0$ is a scalar hyperparameter introduced to control the trade-off between utility and fairness. As can be seen, in the new formulation, the dataset weights are updated by taking the utility loss computed over the clean-labeled $D_{dem}$ into account, and this gives us robustness. To accommodate for this change in Algorithm~\ref{alg:biFair}, we only need to add $L_{u_{dem}}$ to line 13, and consequently compute the gradient over $L_{f} + \lambda\cdot L_{u_{dem}}$ in line 14.

\begin{Ualgorithm}
  \caption{BiFair with automatic differentiation for supervised classification. For each outer iteration, we find an approximate solution to the inner problem (lines 3-9). Then, we compute  the fairness loss $L_f$ on the demographically labeled portion of the data (lines 10-13). Finally, we update the weights of the training datasets (lines 14-15). Note that, we use demographic features only on line 13. Also, computing  $\nabla_w L_{f}$ requires us to maintain the computation graph of the inner-loop. This computation graph is only freed at line 15, after we compute $\nabla_w L_{f}.$}
  \label{alg:biFair}
\DontPrintSemicolon
\SetKwInOut{Input}{Input}
\SetKwInOut{Output}{Output}
\Input{Training dataset $D_{tr}$ with demographically labeled portion denoted as $D_{dem} \subseteq D_{tr}$, and corresponding batch sizes $B_{tr}$, and $B_{dem}$, number of outer iterations $T_{out}$, and number of inner iterations \emph{per} outer iteration $T_{in}$ }
\Output{Trained model M$_\theta$ parameterized by $\theta$}
    
    Initialize $\theta$ and $w$ randomly\;
    \For{$t_{out} \gets 1$ \KwTo $T_{out}$}{
        \For{$t_{in} \gets 1$ \KwTo $T_{in}$}{
        $X_{tr}, y_{tr}, \gets \text{SampleMiniBatch}(D_{tr}, B_{tr})$\;
        $\hat{y}_{tr} \gets \text{ForwardPass}(M_{\theta}, X_{tr})$\;
        $L_{u} \gets \frac{1}{B_{tr}} \sum_i w^{(i)}\cdot L_u(\hat{y}_{tr}^{(i)}, y_{tr}^{(i)})$\;
        $\nabla_\theta L_{u} \gets \text{BackProp}(L_{u}, \theta)$\;
        $\theta \gets \text{OptimizerUpdate}(\theta, \nabla_\theta L_{u})$ \tcp*[h]{SGD, Adam etc.}\;
        }
     $X_{dem}, y_{dem}, s_{dem} \gets \text{SampleMiniBatch}(D_{dem}, B_{dem})$ \tcp*[h]{Demographic features are retrieved here}\;
     $\hat{y}_{dem} \gets \text{ForwardPass}(M_{\theta}, X_{dem})$\;
     $L_{u_{dem}} \gets \frac{1}{B_{dem}} \sum_i L_u(\hat{y}_{dem}^{(i)}, y_{dem}^{(i)})$\;
     $L_{f} \gets \text{ComputeFairnessLoss}(L_{u_{dem}}, y_{dem}, s_{dem})$\ \tcp*[h]{See Table~\ref{tab:biasMetrics} for what this loss can be}\;
     $\nabla_w L_{f} \gets \text{BackProp}(L_{f}, w)$\;
     $w \gets \text{OptimizerUpdate}(w, \nabla_w L_{f})$\;
    }

\end{Ualgorithm}

\section{Experiments}\label{sec:exps}
In this section, we evaluate the performance of the strawman solution, and BiFair via experiments and compare them against several baselines as well. Our implementation is in PyTorch~(\cite{PyTorch}) with Higher library~(\cite{grefenstette2019generalized}), and 
our code and scripts to replicate our results are publicly available at \url{https://github.com/TinfoilHat0/BiFair}.

\subsection{Experimental Setting}
\paragraph{\textbf{Evaluation Metrics:}} In our experiments, we record four metrics of interest. Three of them are bias metrics, AOD, EOD, and SPD, that are listed and defined in Table~\ref{tab:biasMetrics}. In addition to these, we use the balanced accuracy as in~\cite{aif360} to quantify the utility of the models. We choose balanced accuracy over accuracy because most real-world datasets used in fairness research have label-imbalance, including the ones we use, and this makes accuracy a poor metric of choice. Balanced accuracy (BAcc) is equal to the accuracy when there is no label-imbalance, and is defined as follows, 

$$
\text{BAcc} = \frac{\text{TNR} + \text{TPR}}{2},
$$

where TNR and TPR stand for true negative rate, and true positive rate, respectively. For bias metrics, we note that lower values indicate better performance, i.e., the model is fairer when bias values are lower.

\paragraph{\textbf{Baselines:}}
We consider four baselines in our evaluations: \emph{unconstrained training}, 
two parity-based fair training algorithms sampled from the AIF360 toolkit~\cite{aif360}, \emph{Kamiran reweighing} of~\cite{kamiran} and \emph{Prejudice Remover} of~\cite{prjRemover}, and one Rawlsian fair training algorithm, \emph{Adversarially Reweighted Learning} (ARL) of~\cite{lahoti2020fairness}. 
To the best of our knowledge, ARL is the state-of-the-art Rawlsian method. When choosing other fair training algorithms, we have taken benchmarks
presented in AIF 360 toolkit into account, and have chosen the best performing algorithms in its respective category. When the results between two algorithms were too close to call for a clear winner, we have chosen the algorithm that we deem as simpler to implement. We provide a brief description of each baseline below.\\

\emph{Unconstrained Training} simply refers to the  training of model without any fairness constraint, i.e., we are doing empirical loss minimization as usual.

\emph{Kamiran Reweighing} of~\cite{kamiran} is a pre-processing technique that aims to ensure statistical independence between the label and the sensitive attribute by assigning weights to data points. Particularly, the technique first computes the expected probability, $pr_{exp}$, for each combination of sensitive attribute and label under the assumption that the sensitive attribute and the label are independent. Then, they measure the observed probability, $pr_{obs}$, for each combination of sensitive attribute and label in the training dataset. Finally, each data point is assigned the weight $pr_{exp}/pr_{obs}$ based on the value of their sensitive attribute and their label.

\emph{Prejudice Remover} of~\cite{prjRemover} is an in-processing technique that tries to ensure statistical independence between the model's prediction, and the sensitive attribute. To do so, the empirical mutual information between the model's prediction and the sensitive attribute is added as a regularization term in the training objective.

\emph{Adversarially Reweighted Learning (ARL)} of~\cite{lahoti2020fairness}
is an in-processing technique that models the fair training as a min-max game between a \emph{learner} model, and an \emph{adversary} model. The goal of the learner model is to minimize the empirical loss over a weighted dataset where the adversary tries to assign higher weights to samples in which learner model performs worse as indicated by its loss value. Crucially, since the reweighing is done only according to individual loss values of samples, this algorithm does not require any sensitive attribute to be known.

\begin{figure}
\centering
\includegraphics[width=0.8\textwidth]{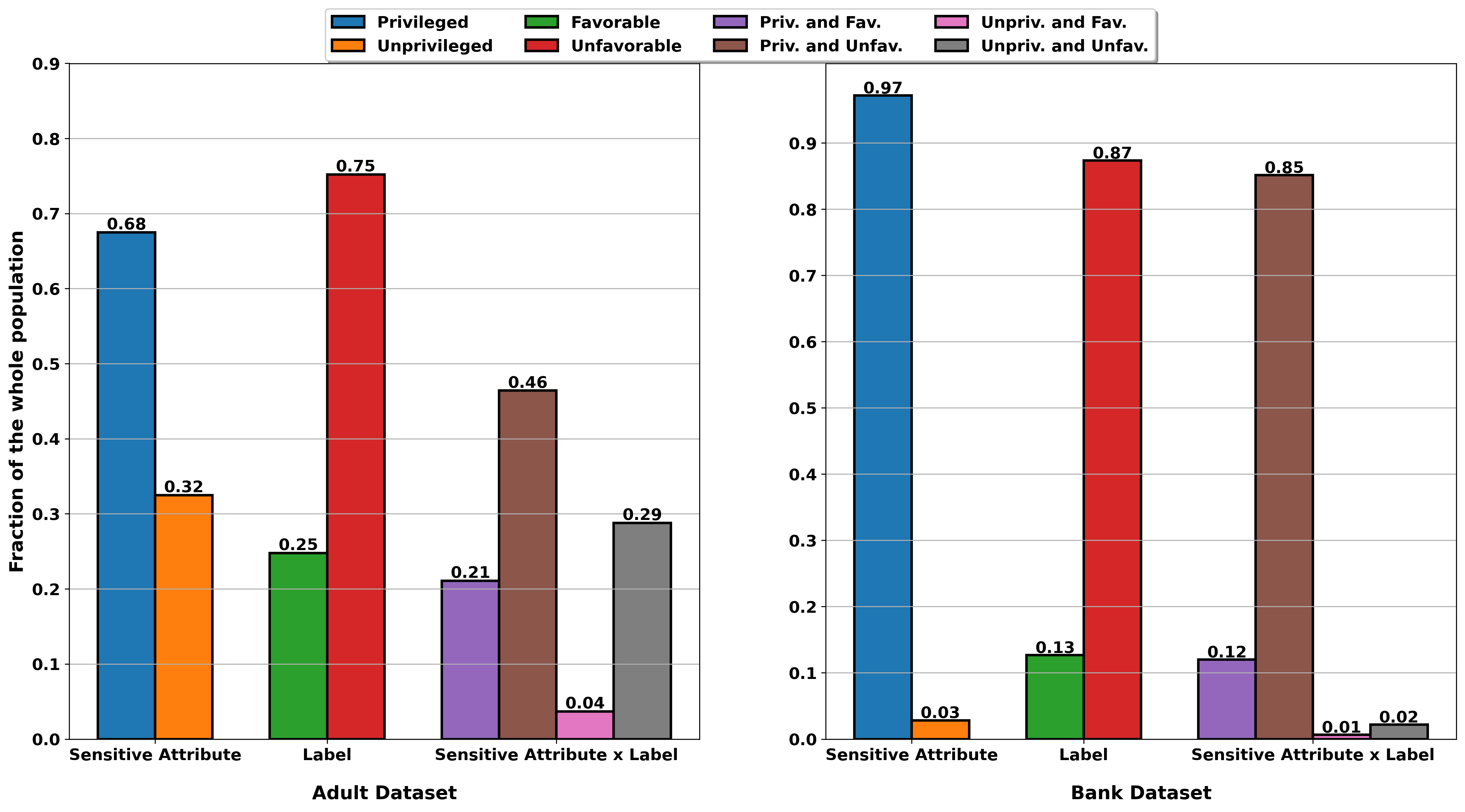}
  \caption{Several statistics regarding fairness for the datasets that we use. First, note that both datasets suffer from severe label imbalance, which justifies our choice of balanced accuracy over accuracy to measure the utility. Further, samples with privileged sensitive attribute value (men for Adult, and old people for Bank) appear more frequently with the favorable label overall. This is not surprising as privileged samples also appear much more frequently than unprivileged samples. However, it is interesting to note that, conditioned on the sensitive attribute, unprivileged samples are more likely to be favorable for the Bank dataset. Concretely, we have $Pr[\text{Label=Favorable} \mid \text{Sensitive Attribute=Privileged}] = 0.12$ and $Pr[\text{Label=Favorable} \mid \text{Sensitive Attribute=Unprivileged}] = 0.23$ in the Bank dataset. We suspect that, due to this, the bias exhibited by the models over the Bank dataset was generally much lower. Consequently, the effect of fair training approaches was more visible in the Adult dataset.}
  \label{fig:dsStats}
\end{figure}

\paragraph{\textbf{Datasets:}} As for the datasets, we use the Adult and Bank datasets from the UCI repository~(\cite{Dua:2019}). In the Adult dataset, the goal is to predict whether a person's income is greater than of \$50k USD a year, the sensitive feature is gender, and men are the privileged. For the Bank dataset, the goal is to predict whether a client will make a deposit subscription or not, the sensitive attribute is age, and old people (older than 25 years) are privileged. We provide some statistics for the datasets we use, and briefly discuss their implications for fairness in Figure~\ref{fig:dsStats}.

\paragraph{\textbf{Setup:}} For each dataset, we ensure a train/validation/test split of $60\%/20\%/20\%$ where we assume model can access the sensitive attributes for only a subset of the training dataset. The sensitive attributes are treated as meta-features, and are not fed to the input layer of models to ensure uniformity across baselines. In all of our experiments, we train a logistic regression model using the Adam optimizer of~\cite{kingma2014adam}, and in the case of BiFair, we also use Adam to update the training dataset weights. Each model is trained until the validation loss stagnates for 5 epochs, i.e., we use early-stopping to decide when to stop training. We report the final measurements done on the test dataset where each measurement is averaged over 10 runs. Measurements are plotted as barcharts for the sake of presentation, and exact values are provided in Appendix~\ref{sec:barplotVals}.

\subsection{Experimental Results}
\begin{figure}
\centering
  \begin{subfigure}[b]{1\textwidth}
  \centering
    \includegraphics[width=1\textwidth]{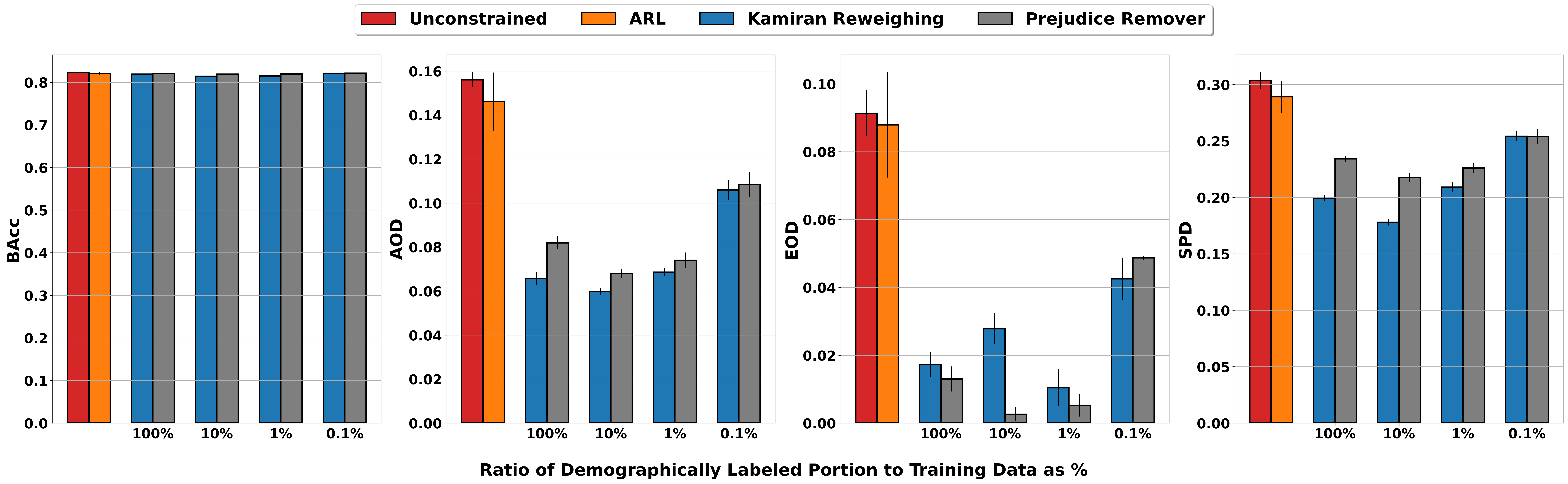}
    \caption{Adult dataset}
  \end{subfigure}
  \begin{subfigure}[b]{1\textwidth}
    \includegraphics[width=1\textwidth]{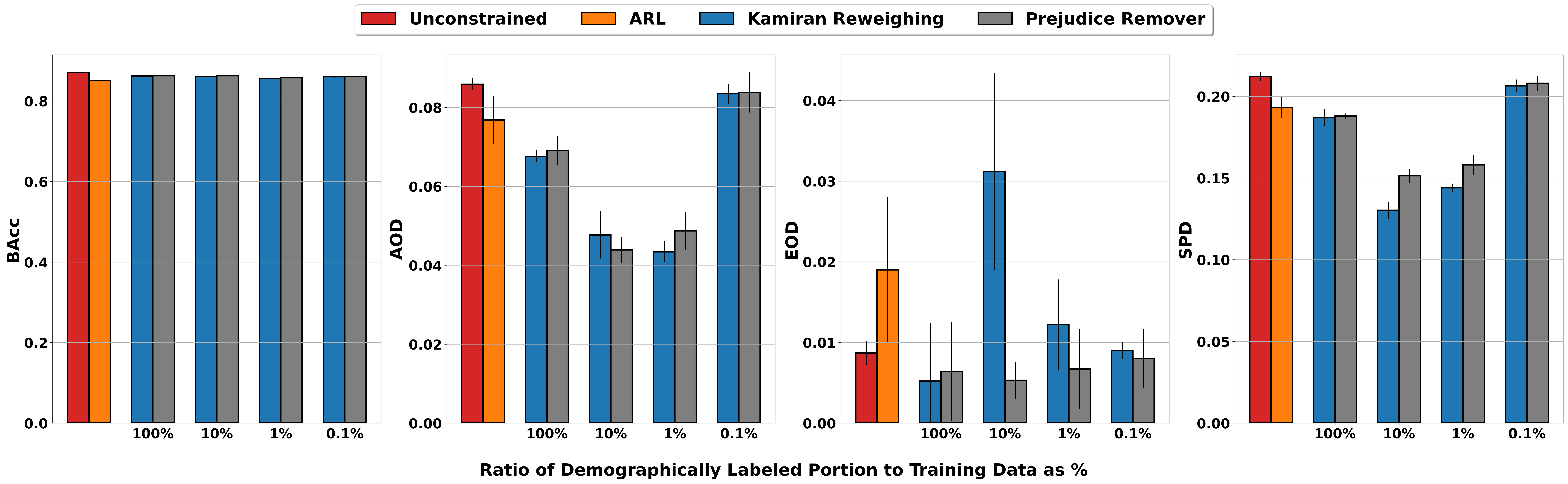}
    \caption{Bank dataset}
  \end{subfigure}
  \caption{The performance of strawman approach against unconstrained training and ARL. 
  First of all, we observe that all methods perform similarly in terms of accuracy.s for the bias metrics, we see all the methods tend to exhibit lower bias compared to the unconstrained training as expected. However, overall results indicate that ARL is the least effective fair training approach of all. For example, for Adult dataset, we see that the strawman approach outperforms ARL by a visible margin for all bias metrics even with only 0.1\% demographically labeled data ($\approx$ 20 samples). As for Bank dataset, ARL tends to perform slightly better than the strawman approach for AOD and SPD metrics for 0.1\% case, and worse for every other case. Finally, it is worth looking at how the strawman solution scales with the size of demographically labeled data. Across all metrics, 0.1\% case has higher bias than 100\% case. With too little demographically labeled data, we cannot predict demographic attributes too accurately as expected.
  However, we observe that, as we move from 100\% to 0.1\%, bias values gets lower in some cases. We think that this happens because some amount of noise in the demographic attributes acts as a regularizer, and results in better generalization performance for bias metrics. This is especially visible in Bank dataset if we compare 100\% case and 10\%-1\% case for AOD.}
  \label{fig:strawman}
\end{figure}

\paragraph{\textbf{Strawman vs. ARL:}} We first illustrate the performance of our strawman approach against unconstrained training and ARL. To do so, we adapt Kamiran Reweighing, and Prejudice Remover to limited demographics setting by imputing unknown demographic attributes as described in Section~\ref{subsec:strawman}. The results are presented and discussed in detail in Figure~\ref{fig:strawman}. In general, we observe that the strawman solution exhibits considerably less bias than ARL even when the size of the demographically labeled data is as small as 1\%-0.1\% of the training data.

\begin{figure}
  \begin{subfigure}{1\textwidth}
  \centering
    \includegraphics[width=1.0\textwidth]{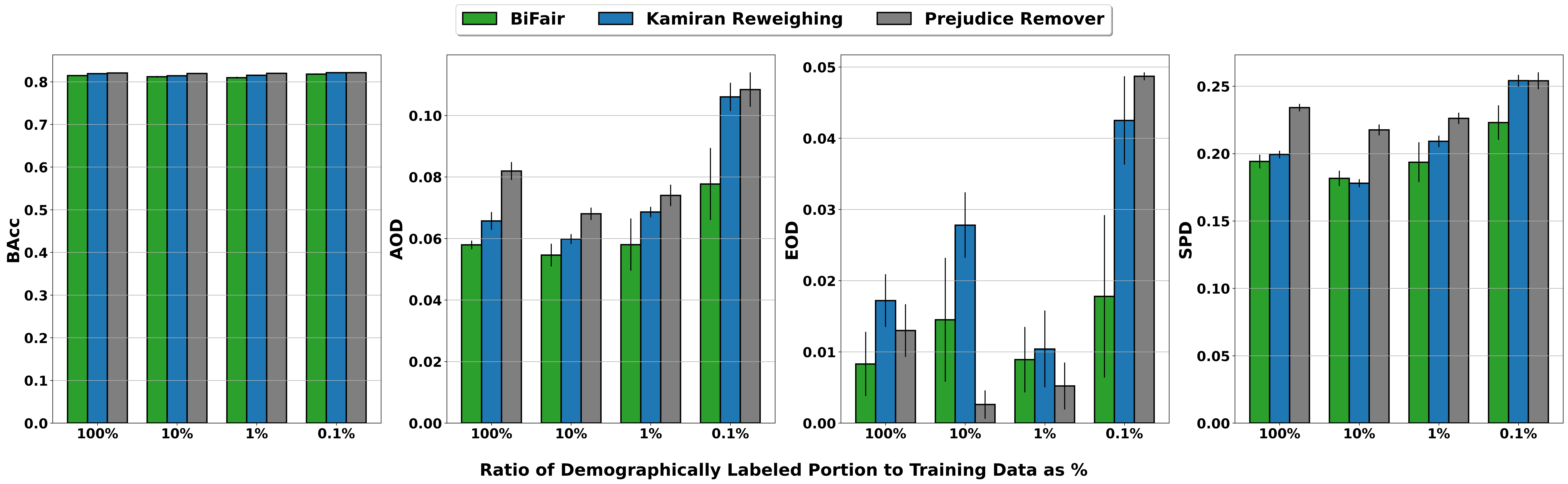}
    \caption{Adult dataset}
  \end{subfigure}
  \begin{subfigure}{1\textwidth}
  \centering
    \includegraphics[width=1.0\textwidth]{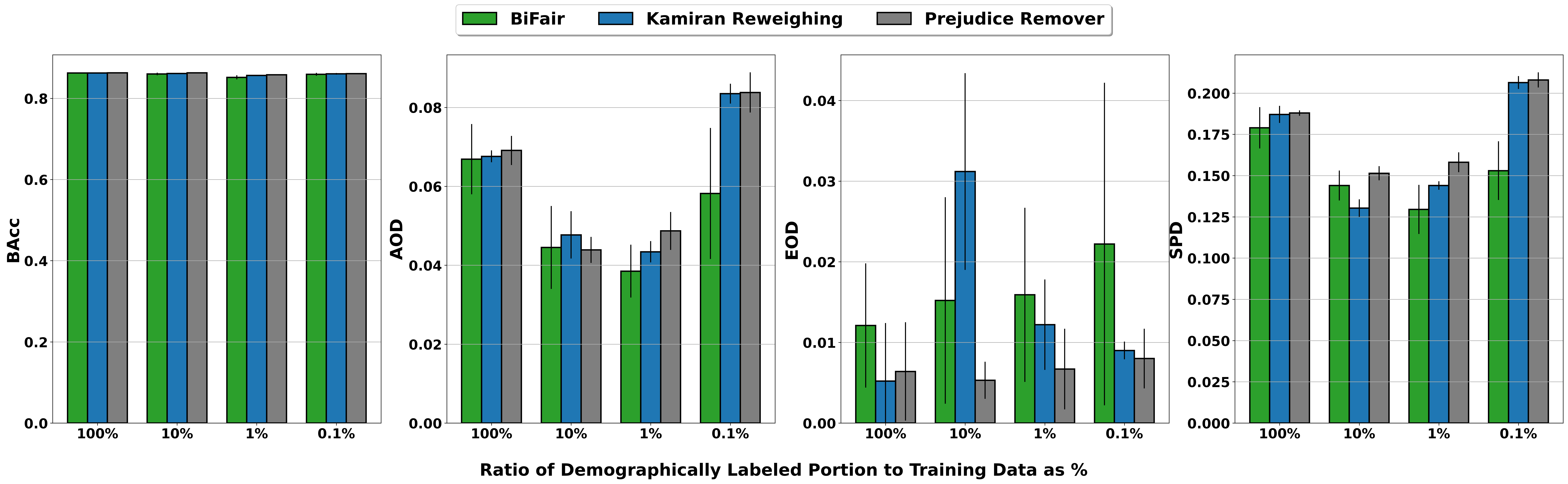}
    \caption{Bank dataset}
  \end{subfigure}
  \caption{The performance of BiFair against the strawman approach.  Again, we see all the approaches tend to perform similar with respect to the accuracy. However, we see that, BiFair tends to degrade more gracefully compared to the strawman approach, and tends to exhibit better fairness as the demographically labeled data gets smaller. This is especially visible for the 0.1\% case. For Adult dataset, it outperforms the strawman approach across all the bias metrics, and for Bank dataset, it is only worse for EOD. We suspect this happens because the Bank datasets exhibits high AOD and low EOD by default (see results of Unconstrained in Figure~\ref{fig:strawman} for Bank). So, perhaps there is a tug-of-war between AOD and EOD metrics for this dataset, and BiFair increases EOD slightly as it brings down AOD.}
  \label{fig:biFair}
\end{figure}

\paragraph{\textbf{BiFair vs. Strawman:}}
Now that we have seen strawman approach can outperform ARL with little demographically labeled data, we compare it with BiFair. For BiFair, we use the AOD loss (see Table~\ref{tab:biasMetrics}) as its fairness loss. This is because AOD encompasses EOD, and unlike SPD, it takes ground-truth labels of data points into account. So, it does not necessary lead to a drop in the utility of the model. We plot and discuss the results in Figure~\ref{fig:biFair}. In brief, the plots suggest that the performance of BiFair degrades more gracefully as the size of the demographically labeled portion gets smaller. Consequently, BiFair tends to outperform the strawman approach especially when the size of the demographically labeled portion is small. To highlight this better, we conducted additional experiments for cases where the demographically labeled portion is less than 1\% of the training data, and provide them in Appendix (Figure~\ref{fig:biFairExtra}).

\begin{figure}
  \begin{subfigure}{1\textwidth}
    \includegraphics[width=1.0\textwidth]{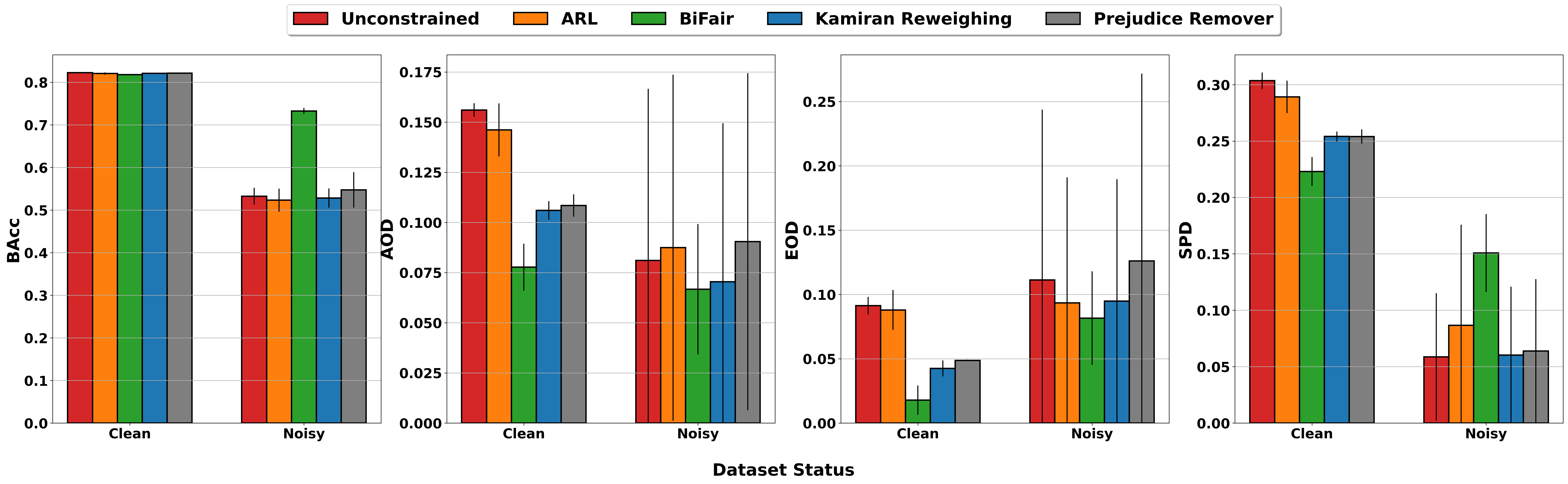}
    \caption{Adult dataset}
  \end{subfigure}
  \begin{subfigure}{1\textwidth}
    \includegraphics[width=1.0\textwidth]{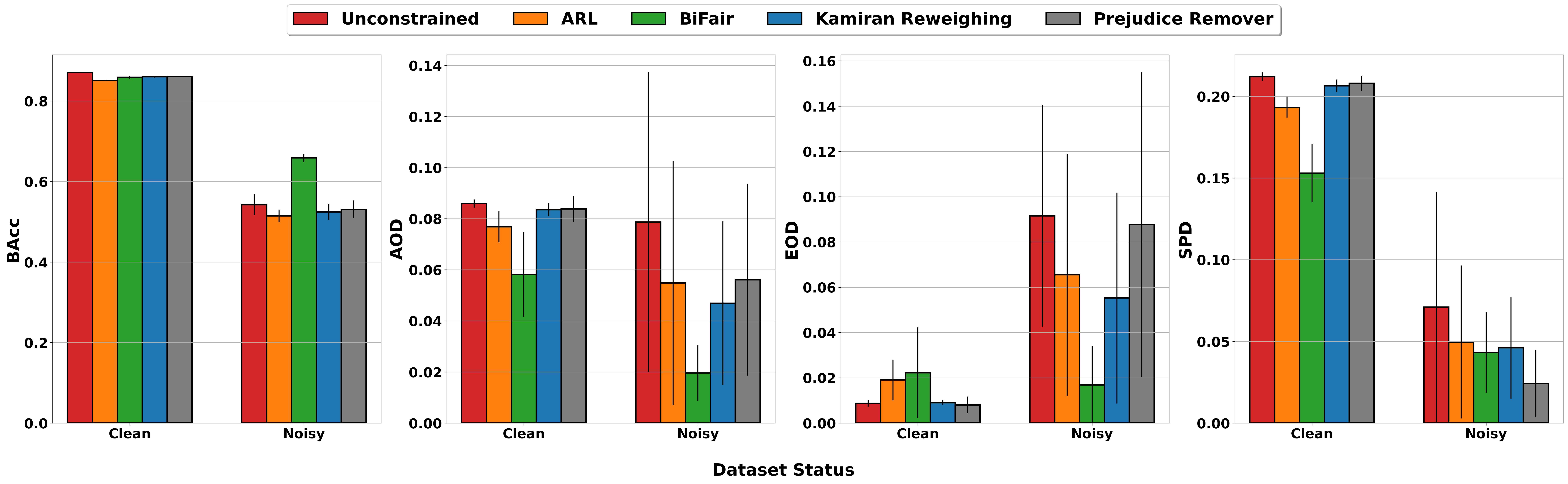}
    \caption{Bank dataset}
  \end{subfigure}
  \caption{Comparison of algorithms under clean, and noisy label settings. As is seen, every other algorithm other than BiFair is susceptible to label noise. In the noisy setting, the accuracy of every algorithm other than BiFair is merely above 50\%. Briefly put, they fail to provide any utility. On the other hand, BiFair maintains much better accuracy with low bias values.}
  
  \label{fig:noisy}
\end{figure}

\paragraph{\textbf{Fairness under Noisy Labels:}}
As we discussed in Section~\ref{subsec:robustness}, we can trivially extend BiFair, and make it robust to label noise assuming our demographically labeled portion  has clean labels (see Equation~\ref{eqn:bifairRobust}). To highlight this property of BiFair, we use the following setting: we first ensure 0.1\% of the training has known demographics, and clean class-labels. For the rest of the data, we flip the class labels with 1/2 probability. The results are presented and discussed in Figure~\ref{fig:noisy}. As can be observed, every other algorithm than BiFair fails to provide any utility, i.e., their accuracies are about 50\%. Meanwhile, BiFair achieves a much better accuracy, while still reducing the bias considerably.

\section{Discussion}
\label{sec:discuss}
We now discuss some aspects of our results that we find interesting, and suggest some avenues for future work. 
First, when evaluating our strawman approach, we have observed that the bias values of models can sometimes get lower as the size of demographically labeled portion shrinks.
Given our prediction accuracy for demographic attributes strictly decreases, as the size of demographically labeled portion decreases, this suggests to us that, some amount of noise on the demographic attributes might act as a regularization effect. Consequently, a carefully-controlled noise amount on demographic attributes might perhaps be used to train fairer models. Therefore, we believe looking at the relationship between noisy demographic attributes, and bias can be an interesting avenue of research. 

As for BiFair, it is worth to briefly discuss the performance overhead, First, note that, we have to maintain the computation graph of the inner optimization (lines 3-9 in Algorithm~\ref{alg:biFair}) until gradient of the outer optimization is computed (line 16 in Algorithm~\ref{alg:biFair}). Consequently, the memory usage of our algorithm scales linearly with $T_{in}$. Although we have observed that we get good performance even with small values of $T_{in}$ in our experimental evaluation\footnote{Concretely, all of our results are obtained with $T_{in}=2$. We did not observe a noticeable improvement in performance for higher values of $T_{in}$.}, the high memory requirements could pose a problem for certain datasets, and large models (e.g., ResNet-101 of~\cite{He_2016_CVPR}). One straightforward way to reduce the memory usage is due to the truncated backpropagation method presented in~\cite{shaban2019truncated}. With truncated backpropagation, we maintain the computation graph only for the last few iterations regardless of the value of $T_{in}$. Consequently, this makes memory requirements independent of $T_{in}$. For example, we can take $100$ inner iterations, yet maintain the computation graph only for the last $5$ steps. This is likely to give a better performance than only taking $5$ inner iterations (see experimental analysis of~\cite{shaban2019truncated}). With respect to computation cost, we see that our algorithm does a forward-backward pass for each inner iteration (line 5 and 7), and then another forward-backward pass at the outer level (line 11 and 15), per iteration. Consequently, this induces a computation overhead factor of  $T_{in}$ + 1 over regular training, which does a single forward-backward pass per iteration. Approaches such as implicit gradients of~\cite{rajeswaran2019meta} might be used to reduce the extra computation cost. \\

Finally, it is worth to note that, although we focused on supervised classification in this work, one can trivially adapt our main formulation given in Equation~\ref{eqn:bifair} to other tasks, such as regression. So, it might be interesting to quantify the performance of our approaches for other tasks under the limited demographics setting.

\section{Conclusion}
\label{sec:conclusion}
We briefly recap our main results before concluding the paper. First, rather surprisingly, we have demonstrated that even a straightforward strawman solution can adapt existing fair training algorithms to limited demographics setting with rather good performance. This implies that industry practitioners who have limited access to demographic attributes can adapt their existing pipelines and algorithms to this setting rather easily.
Second, we have developed a novel algorithm, named BiFair, that is particularly suited to limited demographics setting. Through experiments, we showed that BiFair scales more gracefully as the size of the demographically labeled portion gets smaller, and overall, it tends to exhibit lower bias than the strawman solution.
Further, we have expanded BiFair to make it robust to noisy labels, and illustrated that it can provide both good fairness, and good utility under heavy label noise in the limited demographics setting. In general, we emphasize our main formulation presented in Equation~\ref{eqn:bifair} is quite flexible, and it accommodate for multiple training objectives with little modification as we have shown for fairness and robustness.
In summary, we have developed and evaluated approaches to train fair models in a setting where we know demographic/sensitive attributes for only a subset of the data at hand. To the best of our knowledge, this is the first work to consider such setting in the context of fairness research. Overall, we believe this is a well-motivated setting, and corresponds to what most industry practitioners face in real world. So, we hope that our work will motivate researchers to consider and analyze this particular setting further.

\clearpage
\bibliography{iclr2022_conference}
\bibliographystyle{iclr2022_conference}
\clearpage

\appendix
\section{Additional Experiments for BiFair}
\begin{figure}[h]
  \begin{subfigure}{1\textwidth}
    \includegraphics[width=1\textwidth]{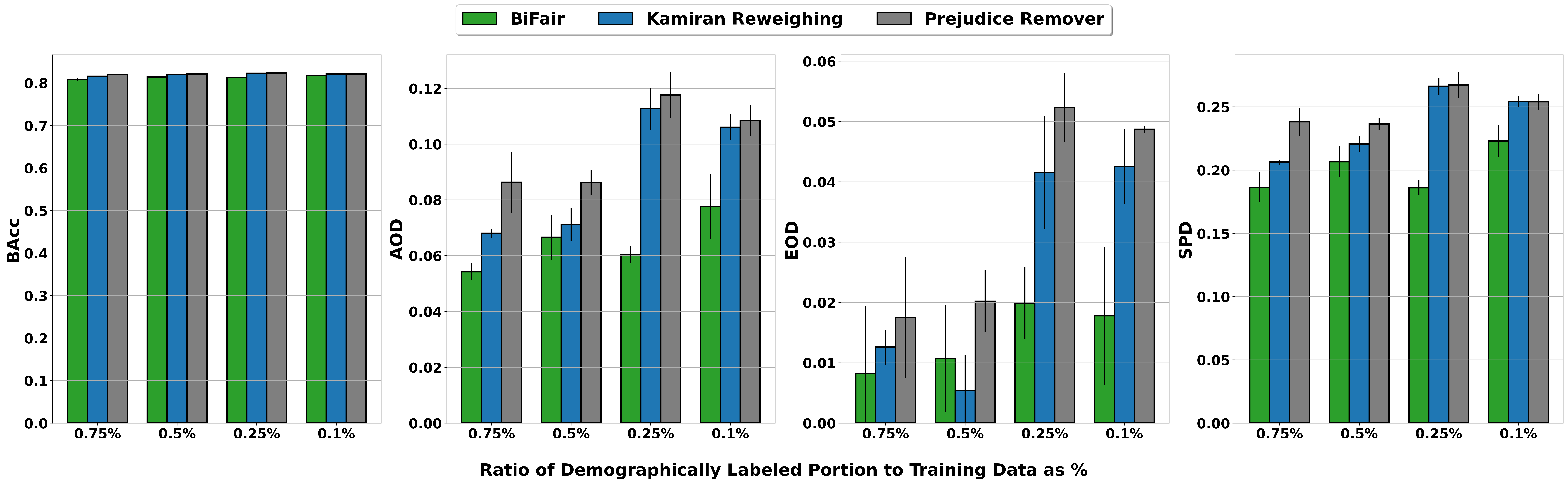}
    \caption{Adult dataset}
  \end{subfigure}
  \begin{subfigure}{1\textwidth}
    \includegraphics[width=1\textwidth]{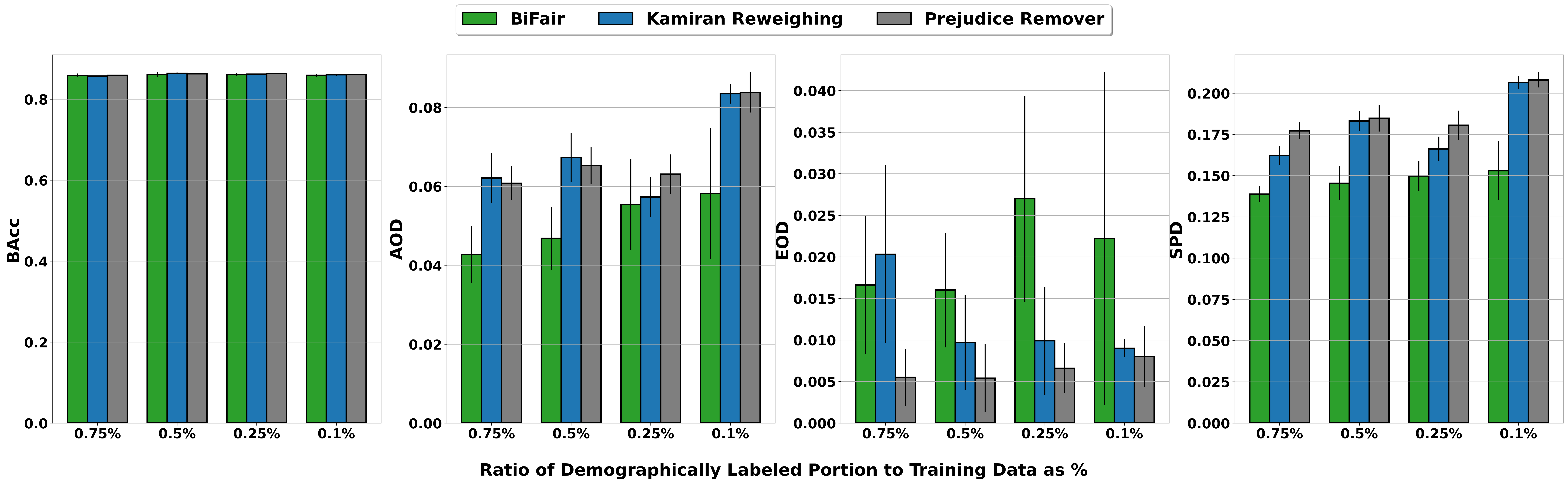}
    \caption{Bank dataset}
  \end{subfigure}
  \caption{The performance of BiFair against the strawman approach when the demographically labeled portion is less than 1\% of the training data. As is seen, BiFair is better across all metrics for 0.25\% and 0.1\% case for the Adult dataset. For the Bank dataset, it is better for AOD and SPD, and worse for EOD (which we have speculated why in Figure~\ref{fig:biFair}).}
  \label{fig:biFairExtra}
\end{figure}

\section{Barchart Values}
\label{sec:barplotVals}
\begin{table}
\caption{Values that are used to generate Figure~\ref{fig:strawman}, Figure~\ref{fig:biFair}, clean dataset part of Figure~\ref{fig:noisy}, and Figure~\ref{fig:biFairExtra}.}
\vspace{-0.2cm}
\centering
    \begin{subtable}{1\textwidth}
      \centering
        \resizebox{0.95\linewidth}{!}{%
        \begin{tabular}{L L L L L L}
        \toprule
           \text{Algorithm}  &\text{Demographic Portion (\%)} &\text{BAcc} &\text{AOD} &\text{EOD} &\text{SPD}\\
        \midrule
    \text{Unconstrained} &- &0.822 \pm 0.0004 &0.156 \pm 0.0034 &0.091 \pm 0.0068 &0.303 \pm 0.0073 \\
    \text{ARL} &- &0.820 \pm 0.0029 &0.146 \pm 0.0132 &0.088 \pm 0.0155 &0.289 \pm 0.0144 \\
    \midrule
    \text{Kamiran Reweighing} &100\% &0.819 \pm 0.0006 &0.066 \pm 0.0029 &0.017 \pm 0.0037 &0.199 \pm 0.0028 \\
    \text{Kamiran Reweighing} &10\% &0.814 \pm 0.0007 &0.060 \pm 0.0016 &0.028 \pm 0.0046 &0.178 \pm 0.0030 \\
    \text{Kamiran Reweighing} &1\% &0.815 \pm 0.0009 &0.069 \pm 0.0017 &0.010 \pm 0.0054 &0.209 \pm 0.0043 \\
    \text{Kamiran Reweighing} &0.75\% &0.816 \pm 0.0007 &0.068 \pm 0.0016 &0.013 \pm 0.0029 &0.206 \pm 0.0020  \\
    \text{Kamiran Reweighing} &0.5\% &0.820 \pm 0.0003 &0.071 \pm 0.0060 &0.005 \pm 0.0059 &0.221 \pm 0.0065  \\
    \text{Kamiran Reweighing} &0.25\% &0.823 \pm 0.0010 &0.113 \pm 0.0075 &0.042 \pm 0.0094 &0.266 \pm 0.0069  \\
    \text{Kamiran Reweighing} &0.1\% &0.821 \pm 0.0006 &0.106 \pm 0.0046 &0.043 \pm 0.0062 &0.254 \pm 0.0044 \\
    \midrule
    \text{Prejudice Remover} &100\% &0.820 \pm 0.0006 &0.082 \pm 0.0029 &0.013 \pm 0.0037 &0.234 \pm 0.0028 \\
    \text{Prejudice Remover} &10\% &0.819 \pm 0.0010 &0.068 \pm 0.0020 &0.003 \pm 0.0020 &0.218 \pm 0.0041 \\
    \text{Prejudice Remover} &1\% &0.820 \pm 0.0005 &0.074 \pm 0.0035 &0.005 \pm 0.0033 &0.226 \pm 0.0042 \\
    \text{Prejudice Remover} &0.75\% &0.820 \pm 0.0007 &0.086 \pm 0.0109 &0.018 \pm 0.0101 &0.238 \pm 0.0111  \\
    \text{Prejudice Remover} &0.5\% &0.821 \pm 0.0008 &0.086 \pm 0.0045 &0.020 \pm 0.0051 &0.236 \pm 0.0049  \\
    \text{Prejudice Remover} &0.25\% &0.823 \pm 0.0017 &0.118 \pm 0.0081 &0.052 \pm 0.0057 &0.267 \pm 0.0100  \\
    \text{Prejudice Remover} &0.1\% &0.821 \pm 0.0008 &0.108 \pm 0.0056 &0.049 \pm 0.0006 &0.254 \pm 0.0063 \\
    \midrule
    \text{BiFair} &100\% &0.814 \pm 0.0015 &0.058 \pm 0.0014 &0.008 \pm 0.0045 &0.194 \pm 0.0051 \\
    \text{BiFair} &10\% &0.812 \pm 0.0020 &0.055 \pm 0.0037 &0.014 \pm 0.0087 &0.182 \pm 0.0057 \\
    \text{BiFair} &1\% &0.809 \pm 0.0021 &0.058 \pm 0.0085 &0.009 \pm 0.0046 &0.194 \pm 0.0148 \\
    \text{BiFair} &0.75\% &0.808 \pm 0.0040 &0.054 \pm 0.0031 &0.008 \pm 0.0112 &0.186 \pm 0.0118  \\
    \text{BiFair} &0.5\% &0.814 \pm 0.0014 &0.067 \pm 0.0081 &0.011 \pm 0.0089 &0.207 \pm 0.0123  \\
    \text{BiFair} &0.25\% &0.813 \pm 0.0014 &0.060 \pm 0.0030 &0.020 \pm 0.0060 &0.186 \pm 0.0059 \\
    \text{BiFair} &0.1\% &0.818 \pm 0.0016 &0.078 \pm 0.0117 &0.018 \pm 0.0114 &0.223 \pm 0.0128 \\
        \bottomrule
        \end{tabular}%
        }
        \caption{\small{Adult dataset}}
    \end{subtable}%
    
    \begin{subtable}{1\textwidth}
      \centering
        \resizebox{0.95\linewidth}{!}{%
        \begin{tabular}{L L L L L L}
        \toprule
           \text{Algorithm}  &\text{Demographic Portion (\%)} &\text{BAcc} &\text{AOD} &\text{EOD} &\text{SPD}\\
        \midrule
    \text{Unconstrained} &-&0.871 \pm 0.0005 &0.086 \pm 0.0016 &0.009 \pm 0.0015 &0.212 \pm 0.0026  \\
    \text{ARL} &- &0.851 \pm 0.0018 &0.077 \pm 0.0061 &0.019 \pm 0.0090 &0.193 \pm 0.0062  \\
    \midrule
    \text{Kamiran Reweighing} &100\% &0.862 \pm 0.0010 &0.068 \pm 0.0015 &0.005 \pm 0.0072 &0.187 \pm 0.0052  \\
    \text{Kamiran Reweighing} &10\% &0.861 \pm 0.0010 &0.048 \pm 0.0060 &0.031 \pm 0.0122 &0.130 \pm 0.0053  \\
    \text{Kamiran Reweighing} &1\% &0.856 \pm 0.0015 &0.043 \pm 0.0027 &0.012 \pm 0.0056 &0.144 \pm 0.0026  \\
    \text{Kamiran Reweighing} &0.75\% &0.857 \pm 0.0012 &0.062 \pm 0.0064 &0.020 \pm 0.0107 &0.162 \pm 0.0057   \\
    \text{Kamiran Reweighing} &0.5\% &0.864 \pm 0.0019 &0.067 \pm 0.0062 &0.010 \pm 0.0057 &0.183 \pm 0.0061   \\
    \text{Kamiran Reweighing} &0.25\% &0.862 \pm 0.0013 &0.057 \pm 0.0051 &0.010 \pm 0.0065 &0.166 \pm 0.0075   \\
    \text{Kamiran Reweighing} &0.1\% &0.860 \pm 0.0019 &0.083 \pm 0.0025 &0.009 \pm 0.0011 &0.206 \pm 0.0039  \\
    \midrule
    \text{Prejudice Remover} &100\% &0.863 \pm 0.0010 &0.069 \pm 0.0037 &0.006 \pm 0.0061 &0.188 \pm 0.0016  \\
    \text{Prejudice Remover} &10\% &0.863 \pm 0.0011 &0.044 \pm 0.0033 &0.005 \pm 0.0023 &0.151 \pm 0.0043  \\
    \text{Prejudice Remover} &1\% &0.858 \pm 0.0010 &0.049 \pm 0.0048 &0.007 \pm 0.0050 &0.158 \pm 0.0060  \\
    \text{Prejudice Remover} &0.75\% &0.859 \pm 0.0009 &0.061 \pm 0.0043 &0.006 \pm 0.0034 &0.177 \pm 0.0051   \\
    \text{Prejudice Remover} &0.5\% &0.863 \pm 0.0009 &0.065 \pm 0.0047 &0.005 \pm 0.0041 &0.185 \pm 0.0081   \\
    \text{Prejudice Remover} &0.25\% &0.863 \pm 0.0012 &0.063 \pm 0.0050 &0.007 \pm 0.0030 &0.181 \pm 0.0088   \\
    \text{Prejudice Remover} &0.1\% &0.861 \pm 0.0016 &0.084 \pm 0.0051 &0.008 \pm 0.0037 &0.208 \pm 0.0046  \\
    \midrule
    \text{BiFair} &100\% &0.862 \pm 0.0016 &0.067 \pm 0.0089 &0.012 \pm 0.0077 &0.179 \pm 0.0125  \\
    \text{BiFair} &10\% &0.860 \pm 0.0031 &0.045 \pm 0.0105 &0.015 \pm 0.0128 &0.144 \pm 0.0091 \\
    \text{BiFair} &1\% &0.852 \pm 0.0051 &0.038 \pm 0.0067 &0.016 \pm 0.0108 &0.130 \pm 0.0149  \\
    \text{BiFair} &0.75\% &0.859 \pm 0.0046 &0.043 \pm 0.0073 &0.017 \pm 0.0083 &0.139 \pm 0.0048 \\
    \text{BiFair} &0.5\% &0.861 \pm 0.0057 &0.047 \pm 0.0080 &0.016 \pm 0.0069 &0.145 \pm 0.0102   \\
    \text{BiFair} &0.25\% &0.861 \pm 0.0039 &0.055 \pm 0.0115 &0.027 \pm 0.0124 &0.150 \pm 0.0091   \\
    \text{BiFair} &0.1\% &0.859 \pm 0.0037 &0.058 \pm 0.0166 &0.022 \pm 0.0200 &0.153 \pm 0.0178  \\
        \bottomrule
        \end{tabular}%
        }
        \caption{Bank dataset}\label{tab:1a}
    \end{subtable}%
\end{table}

\begin{table}
\caption{Values that are used to generate noisy dataset part of Figure~\ref{fig:noisy}. Note that, for this setting, every algorithm has access to a clean-labeled subset of the training data whose size is of 0.1\%, and whose demographic attributes are known.}
\centering
    \begin{subtable}{1\textwidth}
      \centering
        \resizebox{0.95\linewidth}{!}{%
        \begin{tabular}{L L L L L}
        \toprule
           \text{Algorithm}  &\text{BAcc} &\text{AOD} &\text{EOD} &\text{SPD}\\
        \midrule
    \text{Unconstrained} &0.532 \pm 0.0197 &0.081 \pm 0.0856 &0.111 \pm 0.1325 &0.059 \pm 0.0565  \\
    \text{ARL }&0.523 \pm 0.0270 &0.087 \pm 0.0863 &0.093 \pm 0.0976 &0.087 \pm 0.0892 \\
    \text{Kamiran Reweighing} &0.528 \pm 0.0225 &0.070 \pm 0.0790 &0.095 \pm 0.0948 &0.060 \pm 0.0608  \\
    \text{Prejudice Remover} &0.547 \pm 0.0417 &0.090 \pm 0.0840 &0.126 \pm 0.1457 &0.064 \pm 0.0638  \\
    \text{BiFair} &0.733 \pm 0.0069 &0.067 \pm 0.0325 &0.082 \pm 0.0364 &0.151 \pm 0.0345 \\
        \bottomrule
        \end{tabular}%
        }
        \caption{Adult dataset}
    \end{subtable}%
    \\
    \begin{subtable}{1\textwidth}
      \centering
        \resizebox{0.95\linewidth}{!}{%
        \begin{tabular}{L L L L L}
        \toprule
           \text{Algorithm}  &\text{BAcc} &\text{AOD} &\text{EOD} &\text{SPD}\\
        \midrule
    \text{Unconstrained} &0.542 \pm 0.0258 &0.079 \pm 0.0586 &0.091 \pm 0.0490 &0.071 \pm 0.0703  \\
    \text{ARL} &0.515 \pm 0.0156 &0.055 \pm 0.0478 &0.066 \pm 0.0534 &0.050 \pm 0.0468   \\
    \text{Kamiran Reweighing} &0.524 \pm 0.0201 &0.047 \pm 0.0320 &0.055 \pm 0.0466 &0.046 \pm 0.0312   \\
    \text{Prejudice Remover} &0.531 \pm 0.0221 &0.056 \pm 0.0375 &0.088 \pm 0.0672 &0.024 \pm 0.0207   \\
    \text{BiFair} &0.659 \pm 0.0097 &0.020 \pm 0.0108 &0.017 \pm 0.0172 &0.043 \pm 0.0246  \\
        \bottomrule
        \end{tabular}%
        }
        \caption{Bank dataset}\label{tab:1a}
    \end{subtable}%
\end{table}

\end{document}